\documentclass{article} 
\usepackage{iclr2016_conference,times}
\usepackage{url}

\usepackage{times}

\usepackage{graphicx}
\usepackage{caption}
\usepackage{subcaption}

\usepackage{amsmath}
\usepackage{amsfonts}
\usepackage{amssymb}
\usepackage{amsthm}

\usepackage[utf8]{inputenc}
\usepackage{amsthm}
\usepackage{amsmath}
\usepackage{amsfonts}
\usepackage{amssymb}
\usepackage{dsfont}
\usepackage{color}
\allowdisplaybreaks
\newcommand{\R}{\mathbb{R}}
\newcommand{\N}{\mathcal{N}}
\newcommand{\cL}{\mathcal{L}}

\newcommand{\svert}{~|~}
\newcommand{\td}{\text{d}}
\newcommand{\f}{\mathbf{f}}
\newcommand{\x}{\mathbf{x}}
\newcommand{\Bb}{\mathbf{b}}
\newcommand{\sBb}{\mathtt{z}}

\newcommand{\y}{\mathbf{y}}

\newcommand{\bk}{\mathbf{k}}

\newcommand{\W}{\mathbf{W}}

\newcommand{\X}{\mathbf{X}}
\newcommand{\Y}{\mathbf{Y}}
\newcommand{\F}{\mathbf{F}}
\newcommand{\I}{\mathbf{I}}
\newcommand{\M}{\mathbf{M}}

\newcommand{\bz}{\mathbf{0}}

\newcommand{\bo}{\text{\boldmath$\omega$}}

\newcommand{\K}{\mathbf{K}}

\newcommand{\diag}{\text{diag}}
\newcommand{\KL}{\text{KL}}

\newcommand{\weightdecay}{\lambda}

\theoremstyle{definition}

\usepackage{hyperref}

\title{Bayesian Convolutional Neural Networks with Bernoulli Approximate Variational \\ Inference}

\author{Yarin Gal \& Zoubin Ghahramani \\
University of Cambridge\\
\texttt{\{yg279,zg201\}@cam.ac.uk} 
}

\begin{document}

\maketitle

\begin{abstract} 
Convolutional neural networks (CNNs) work well on large datasets. But labelled data is hard to collect, and in some applications larger amounts of data are not available. The problem then is how to use CNNs with small data -- as CNNs overfit quickly. We present an efficient Bayesian CNN, offering better robustness to over-fitting on small data than traditional approaches. This is by placing a probability distribution over the CNN's \textit{kernels}. We approximate our model's intractable posterior with Bernoulli variational distributions, requiring no additional model parameters. 

On the theoretical side, we cast dropout network training as approximate inference in Bayesian neural networks. This allows us to implement our model using existing tools in deep learning with no increase in time complexity, while highlighting a negative result in the field. We show a considerable improvement in classification accuracy compared to standard techniques and improve on published state-of-the-art results for CIFAR-10.
\end{abstract} 

\section{Introduction}

Convolutional neural networks (CNNs), popular deep learning tools for image processing, can solve tasks that until recently were considered to lay beyond our reach  \citep{krizhevsky2012imagenet, szegedy2014going}. However CNNs require huge amounts of data for regularisation and quickly over-fit on small data. In contrast Bayesian neural networks (NNs) are robust to over-fitting, offer uncertainty estimates, and can easily learn from small datasets. First developed in the '90s and studied extensively since then \citep{mackay1992practical, neal1995bayesian}, Bayesian NNs offer a probabilistic interpretation of deep learning models by inferring distributions over the models' weights. 
However, modelling a distribution over the kernels (also known as filters) of a CNN has never been attempted successfully before, perhaps because of the vast number of parameters and extremely large models commonly used in practical applications. 

Even with a small number of parameters, inferring model posterior in a Bayesian NN is a difficult task. Approximations to the model posterior are often used instead, with variational inference being a popular approach. In this approach one would model the posterior using a simple \textit{variational} distribution such as a Gaussian, and try to fit the distribution's parameters to be as close as possible to the true posterior. This is done by minimising the Kullback-Leibler divergence from the true posterior. Many have followed this approach in the past for standard NN models \citep{hinton1993keeping,barber1998ensemble,graves2011practical,blundell2015weight}.
But the variational approach used to approximate the posterior in Bayesian NNs can be fairly computationally expensive -- the use of Gaussian approximating distributions increases the number of model parameters considerably, without increasing model capacity by much. \citet{blundell2015weight} for example use Gaussian distributions for Bayesian NN posterior approximation and have doubled the number of model parameters, yet report the same predictive performance as traditional approaches using dropout. This makes the approach unsuitable for use with CNNs as the increase in the number of parameters is too costly.
Instead, here we use \textit{Bernoulli approximating variational distributions}.
The use of Bernoulli variables requires no additional parameters for the approximate posteriors, and allows us to obtain a computationally efficient Bayesian CNN implementation. 

Perhaps surprisingly, we can implement our model using existing tools in the field.
\citet{Gal2015Dropout} have recently shown that dropout in NNs can be interpreted as an approximation to a well known Bayesian model -- the Gaussian process (GP). 
What was not shown, however, is how this relates to Bayesian NNs or to CNNs, and was left for future research \citep[][appendix section 4.2]{Gal2015Dropout}. 
Extending on the work, we show here that dropout networks' training can be cast as approximate Bernoulli variational inference in Bayesian NNs. 
This allows us to use operations such as convolution and pooling in probabilistic models in a principled way. 
The implementation of our Bayesian neural network is thus reduced to performing dropout after every convolution layer at training. This, in effect, approximately integrates over the kernels. 
At test time we evaluate the model output by approximating the predictive posterior -- we average stochastic forward passes through the model -- referred to as Monte Carlo (MC) dropout. 

Our model is implemented by performing dropout after convolution layers. In existing literature, however, dropout is often \textit{not used} after convolution layers. This is because test error suffers, which renders small dataset modelling a difficult task. 
This highlights a \textit{negative result} in the field: the dropout approximation fails with convolutions. 
Our mathematically grounded solution alleviates this problem by interleaving Bayesian techniques into deep learning. 

Following our theoretical insights we propose new practical dropout CNN architectures, mathematically identical to Bayesian CNNs. These models obtain better test accuracy compared to existing approaches in the field with no additional computational cost during training. 
We show that the proposed model reduces over-fitting on small datasets compared to standard techniques. Furthermore, we demonstrate improved results with MC dropout on existing CNN models in the literature. 
We give empirical results assessing the number of MC samples required to improve model performance, and finish with state-of-the-art results on the CIFAR-10 dataset following our insights. 
The main contributions of the paper are thus: 
\begin{enumerate}
\item
Showing that the dropout approximation fails in some network architectures. This extends on the results given in \citep{srivastava2014dropout}. 
This is why dropout is not used with convolutions in practice, and as a result CNNs overfit quickly on small data.
\item
Casting dropout as variational inference in Bayesian neural networks, 
\item
This Bayesian interpretation of dropout allows us to propose the use of MC dropout for convolutions (fixing the problem with a mathematically grounded approach, rather than through trial and error), 
\item
Comparing the resulting techniques empirically.
\end{enumerate}

The paper is structured as follows. In section 2 we briefly review required background. In section 3 we derive results connecting dropout to approximate inference in Bayesian NNs, and explain their relation to the results of \citet{Gal2015Dropout} in section 4.
In section 5 we present the Bayesian CNN as an example Bayesian NN model, taking advantage of convolution operations. Finally, in section 6 we give a thorough experimental evaluation of the proposed model.

\section{Background}

We next review probabilistic modelling and variational inference --- the foundations of our derivations. These are followed by a quick review of dropout and Bayesian NNs.
We will link dropout in NNs to approximate variational inference in Bayesian NNs in the next section.

\subsection{Probabilistic Modelling and Variational Inference}
Given training inputs $\{ \x_1, \hdots, \x_N \}$ and their corresponding outputs $\{\y_1, \hdots, \y_N\}$, in probabilistic modelling we would like to estimate a function $\y = \f(\mathbf{x})$ that is \textit{likely to have generated our outputs}. 
What is a function that is likely to have generated our data? Following the Bayesian approach we would put some \textit{prior} distribution over the space of functions $p(\f)$. This distribution represents our prior belief as to which functions are likely to have generated our data. 
We define a \textit{likelihood} $p(\Y | \f, \X)$ to capture the process in which observations are generated given a specific function.
We then look for the \textit{posterior} distribution over the space of functions given our dataset: $p(\f | \X, \Y)$.
This distribution captures the most likely functions given our observed data.
With it we can predict an output for a new input point $\x^*$ by integrating over all possible functions $\f$,
\begin{align} \label{eq:post}
p(\y^* | \x^*, \X, \Y) = \int p(\y^* | \f^*) p(\f^* | \x^*, \X, \Y) \td \f^*.
\end{align}

Integral \eqref{eq:post} is intractable for many models. 
To approximate it we could condition the model on a finite set of random variables $\bo$. We make a modelling assumption and assume that the model depends on these variables alone, making them into sufficient statistics in our approximate model.

The predictive distribution for a new input point $\x^*$ is then given by 
$$
p(\y^* | \x^*, \X, \Y) = \int p(\y^* | \f^*) p(\f^* | \x^*, \bo) p(\bo | \X, \Y)\ \td \f^* \td \bo.
$$
The distribution $p(\bo | \X, \Y)$ cannot usually be evaluated analytically as well. Instead we define an approximating \textit{variational} distribution $q(\bo)$, whose structure is easy to evaluate.
We would like our approximating distribution to be as close as possible to the posterior distribution obtained from the original model. We thus minimise the Kullback--Leibler (KL) divergence, intuitively a measure of similarity between two distributions: $\KL(q(\bo) ~||~ p(\bo | \X, \Y))$,
resulting in the approximate predictive distribution 
\begin{align} \label{eq:predictive}
q(\y^* | \x^*) = \int p(\y^* | \f^*) p(\f^* | \x^*, \bo) q(\bo) \td \f^* \td \bo.
\end{align}

Minimising the Kullback--Leibler divergence is equivalent to maximising the \textit{log evidence lower bound},
\begin{align}
&\cL_{\text{VI}} := \int q(\bo) p(\F | \X, \bo) \log p(\Y | \F) \td \F \td \bo - \KL(q(\bo) || p(\bo)) \label{eq:L:VI}
\end{align}
with respect to the variational parameters defining $q(\bo)$. This is known as \textit{variational inference}, a standard technique in Bayesian modelling.

\subsection{Dropout}
Let $\widehat{\y}$ be the output of a NN with $L$ layers and a loss function $E(\cdot,\cdot)$ such as the softmax loss or the Euclidean loss (squared loss). We denote by $\W_i$ the NN's weight matrices of dimensions $K_i \times K_{i-1}$, and by $\Bb_i$ the bias vectors of dimensions $K_i$ for each layer $i = 1, ..., L$. 
During NN optimisation a regularisation term is often used.
We often use $L_2$ regularisation weighted by some weight decay $\weightdecay$, resulting in a minimisation objective (often referred to as cost),
\begin{align} \label{eq:L:dropout}
\cL_{\text{dropout}} := \frac{1}{N} \sum_{i=1}^N E(\y_i,\widehat{\y}_i) + \weightdecay \sum_{i=1}^L \big( ||\W_i||^2_2 + ||\Bb_i||^2_2 \big).
\end{align}
With dropout, we sample binary variables for every input point and for every network unit in each layer. Each binary variable takes value 1 with probability $p_i$ for layer $i$. A unit is dropped (i.e.\ its value is set to zero) for a given input if its corresponding binary variable takes value 0. We use the same binary variable values in the backward pass propagating the derivatives to the parameters. 

\subsection{Bayesian Neural Networks}

One defines a Bayesian NN by placing a prior distribution over a NN's weights. Given weight matrices $\W_i$ and bias vectors $\Bb_i$ for layer $i$, we often place standard matrix Gaussian prior distributions over the weight matrices, $p(\W_i)$:
\begin{align*}
\W_i \sim \N(\bz, \I).
\end{align*}
We may assume a point estimate for the bias vectors for simplicity. We denote the random output of a NN with weight random variables $(\W_i)_{i=1}^L$ on input $\x$ by $\widehat{\f} \big( \x, (\W_i)_{i=1}^L \big)$, and in classification tasks often assume a softmax likelihood given the NN's weights:
\begin{align*}
p \big( y | \x, (\W_i)_{i=1}^L \big) = \text{Categorical}\left( \exp( \widehat{\f}) / \sum_{d'} \exp( \widehat{f}_{d'}) \right)
\end{align*}
with $\widehat{\f} = \widehat{\f} \big( \x, (\W_i)_{i=1}^L \big)$ a random variable. Even though Bayesian NNs seem simple, calculating model posterior is a hard task. This is discussed next.

\section{Dropout as Approximate Variational Inference in \\ Bayesian Neural Networks}

We now develop approximate variational inference in Bayesian NNs using Bernoulli approximating variational distributions, and relate this to dropout training. This extends on \citep{Gal2015Dropout} as explained in the next section. 

As before, we are interested in finding the most probable functions that have generated our data. In the Bayesian NN case the functions are defined through the NN weights, and these are our sufficient statistics $\bo = (\W_i)_{i=1}^L$. We are thus interested in the posterior over the weights given our observables $\X, \Y$: $p \big( \bo | \X, \Y \big)$. 
This posterior is not tractable for a Bayesian NN, and we use variational inference to approximate it. 

To relate the approximate inference in our Bayesian NN to dropout training, we define our approximating variational distribution $q(\W_i)$ for every layer $i$ as 
\begin{align}\label{eq:approx-dist}
\W_i &= \M_i \cdot \diag([\sBb_{i,j}]_{j=1}^{K_i}) \\
\sBb_{i,j} &\sim \text{Bernoulli}(p_i) \text{ for } i = 1, ..., L, ~ j = 1, ..., K_{i-1}. \notag
\end{align}
Here $\sBb_{i,j}$ are Bernoulli distributed random variables with some probabilities $p_i$, and $\M_i$ are variational parameters to be optimised.
The $\diag(\cdot)$ operator maps vectors to diagonal matrices whose diagonals are the elements of the vectors. 

The integral in eq.\ \eqref{eq:L:VI} is intractable and cannot be evaluated analytically for our approximating distribution. Instead, we approximate the integral with Monte Carlo integration over $\bo$. This results in an unbiased estimator for $\cL_{VI}$:
\begin{align}
\widehat{\cL}_{\text{VI}} := \sum_{i=1}^N E \big( \y_i, \widehat{\f} ( \x_i, \widehat{\bo}_i ) \big) - \KL(q(\bo) || p(\bo)) && \widehat{\bo}_i \sim q(\bo)
\end{align}
with $E(\cdot, \cdot)$ being the softmax loss (for a softmax likelihood).
Note that sampling from $q(\W_i)$ is identical to performing dropout on layer $i$ in a network whose weights are $(\M_i)_{i=1}^L$.
The binary variable $\sBb_{i,j} = 0$ corresponds to unit $j$ in layer $i-1$ being dropped out as an input to the $i$'th layer. The second term in eq.\ \eqref{eq:L:VI} can be approximated following \citep{Gal2015Dropout}, resulting in the objective eq.\ \eqref{eq:L:dropout}.
Dropout and Bayesian NNs, in effect, result in the same model parameters that best explain the data.

Predictions in this model follow equation \eqref{eq:predictive} replacing the posterior $p \big( \bo| \X, \Y \big)$ with the approximate posterior $q \big( \bo \big)$. We can approximate the integral with Monte Carlo integration:
\begin{align} \label{eq:approx_predictive}
p(y^* | \x^*, \X, \Y) \approx 
\int p \big( y^* | \x^*, \bo \big) q \big( \bo \big) 
\td \bo
\approx \frac{1}{T} \sum_{t=1}^T p ( y^* | \x^*, \widehat{\bo}_t )
\end{align}
with $\widehat{\bo}_t \sim q \big( \bo \big)$. This is referred to as MC dropout.

\section{Relation to Gaussian Processes}
Our work extends on the results of \citep{Gal2015Dropout}, relating dropout to approximate inference in the Gaussian process. 

The Gaussian process (GP) is a powerful tool in statistics that allows us to model distributions over functions \citep{Rasmussen2005Gaussian}. 
For regression for example we would place a joint Gaussian distribution over all function values $\F = [\f_1, ..., \f_N]$, and generate observations from a normal distribution centred at $\F$, 
\begin{align} \label{eq:generative_model_class}
\F \svert \X &\sim \N(\bz, \K(\X, \X)) \\
\y_n \svert \f_n &\sim \N ( \y_n; \f_{n}, \tau^{-2}) \notag
\end{align}
for $n = 1, ..., N$ with a covariance function $\K(\cdot, \cdot)$ and noise precision $\tau$.

\citet{Gal2015Improving} have shown that the Gaussian process can be approximated by defining a Gaussian approximating distribution over the spectral frequencies and their coefficients in a Fourier decomposition of the function $\f$.
\citet{Gal2015Dropout} have extended that work showing that by defining the approximating distribution as in eq.\ \eqref{eq:approx-dist}, the resulting objective function is identical to dropout's objective in \textbf{deep networks}. 
Our extension of the model beyond the Gaussian process allows us to represent convolution operations with a Bayesian interpretation. These do not necessarily have a corresponding GP interpretation, but can be modelled as Bayesian NNs.

\section{Bayesian Convolutional Neural Networks}
A direct result of our theoretical development in the previous sections is that Bernoulli approximate variational inference in Bayesian NNs can be implemented by adding dropout layers after certain weight layers in a network. Implementing our Bayesian neural network thus reduces to performing dropout after every layer with an approximating distribution at training, and evaluating the predictive posterior using eq.\ \eqref{eq:approx_predictive} at test time. In Bayesian NNs often all weight layers are modelled with distributions -- the posterior distribution acts as a regulariser, approximately integrating over the weights. Weight layers with no approximating distributions would often lead to over-fitting. In existing literature, however, dropout is used in CNNs only after inner-product layers -- equivalent to approximately integrating these alone. 
Here we wish to integrate over the kernels of the CNN as well.
Thus implementing a Bayesian CNN we apply dropout after all convolution layers as well as inner-product layers. 

To integrate over the kernels, we reformulate the convolution as a linear operation -- an inner-product to be exact.
Let $\bk_k \in \R^{h \times w \times K_{i-1}}$ for $k = 1, ..., K_i$ be the CNN's kernels with height $h$, width $w$, and $K_{i-1}$ channels in the $i$'th layer. The input to the layer is represented as a 3 dimensional tensor $\x \in \R^{H_{i-1} \times W_{i-1} \times K_{i-1}}$  with height $H_{i-1}$, width $W_{i-1}$, and $K_{i-1}$ channels. Convolving the kernels with the input with a given stride $s$ is equivalent to extracting patches from the input and performing a matrix product: we extract $h \times w \times K_{i-1}$ dimensional patches from the input with stride $s$ and vectorise these. Collecting the vectors in the rows of a matrix we obtain a new representation for our input $\overline{\x} \in \R^{n \times hwK_{i-1}}$ with $n$ patches. The vectorised kernels form the columns of the weight matrix $\W_i \in \R^{hwK_{i-1} \times K_i}$. The convolution operation is then equivalent to the matrix product $\overline{\x} \W_i \in \R^{n \times K_i}$. The columns of the output can be re-arranged to a 3 dimensional tensor $\y \in \R^{H_{i} \times W_{i} \times K_{i}}$ (since $n = H_{i}\times W_{i} $). Pooling can then be seen as a non-linear operation on the matrix $\y$. 
Note that the pooling operation is a non-linearity applied after the linear convolution counterpart to ReLU or Tanh non-linearities in \citep{Gal2015Dropout}.

We place a prior distribution over each kernel and approximately integrate each kernels-patch pair with Bernoulli variational distributions. 
We sample Bernoulli random variables $\sBb_{i,j,n}$ and multiply patch $n$ by the weight matrix 
$\W_i \cdot \diag([\sBb_{i,j,n}]_{j=1}^{K_i}).$
This is equivalent to an approximating distribution modelling each kernel-patch pair with a distinct random variable, tying the means of the random variables over the patches.
This distribution randomly sets kernels to zero for different patches. This is also equivalent to applying dropout for each element in the tensor $\y$ before pooling. Implementing our Bayesian CNN is therefore as simple as using dropout after every convolution layer before pooling.

The standard dropout test time approximation does not perform well when dropout is applied after convolutions -- this is a negative result we identified empirically. We solve this by approximating the predictive distribution following eq.\ \eqref{eq:approx_predictive}, averaging stochastic forward passes through the model at test time (using MC dropout). 
We next assess the model above with an extensive set of experiments studying its properties.

\section{Experiments}

We evaluate the theoretical insights brought above by implementing our Bernoulli Bayesian CNNs using dropout. We show that a considerable improvement in classification performance can be attained with a mathematically principled use of dropout on a variety of tasks, assessing the LeNet network structure \citep{lecun1998gradient} on MNIST \citep{lecun1998mnist} and CIFAR-10  \citep{krizhevsky2009learning} with different settings. 
We then inspect model over-fitting by training the model on small random subsets of the MNIST dataset.
We test various existing model architectures in the literature with MC dropout (eq.\ \eqref{eq:approx_predictive}).
We then empirically evaluate the number of samples needed to obtain an improvement in results. We finish with state-of-the-art results on CIFAR-10 obtained by an almost trivial change of an existing model. All experiments were done using the Caffe framework \citep{jia2014caffe}, 
requiring identical training time to that of standard CNNs,
with the configuration files available online at \url{http://mlg.eng.cam.ac.uk/yarin/}. 

\subsection{Bayesian Convolutional Neural Networks} \label{sect:Bayesian Convolutional Neural Networks}

We show that performing dropout after all convolution and weight layers (our Bayesian CNN implementation) in the LeNet CNN on both the MNIST dataset and CIFAR-10 dataset results in a considerable improvement in test accuracy compared to existing techniques in the literature.

\begin{figure}[b!]
\vspace{-5mm}
	\begin{subfigure}[b]{0.5\textwidth}
		\includegraphics[width=\linewidth]{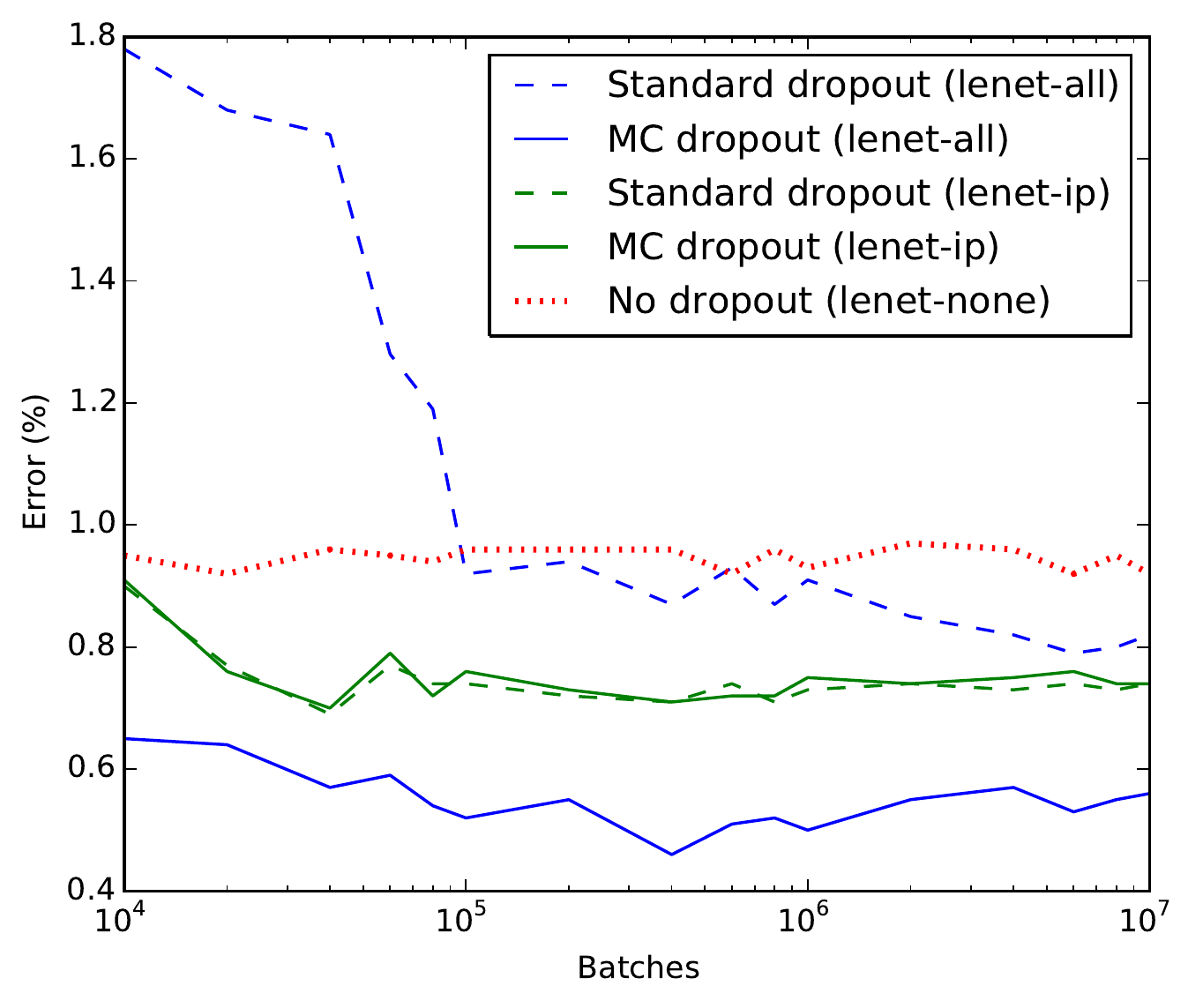}
		\caption{MNIST} \label{fig:bayesian-CNN-mnist}
	\end{subfigure}
	\begin{subfigure}[b]{0.5\textwidth}
		\includegraphics[width=\linewidth]{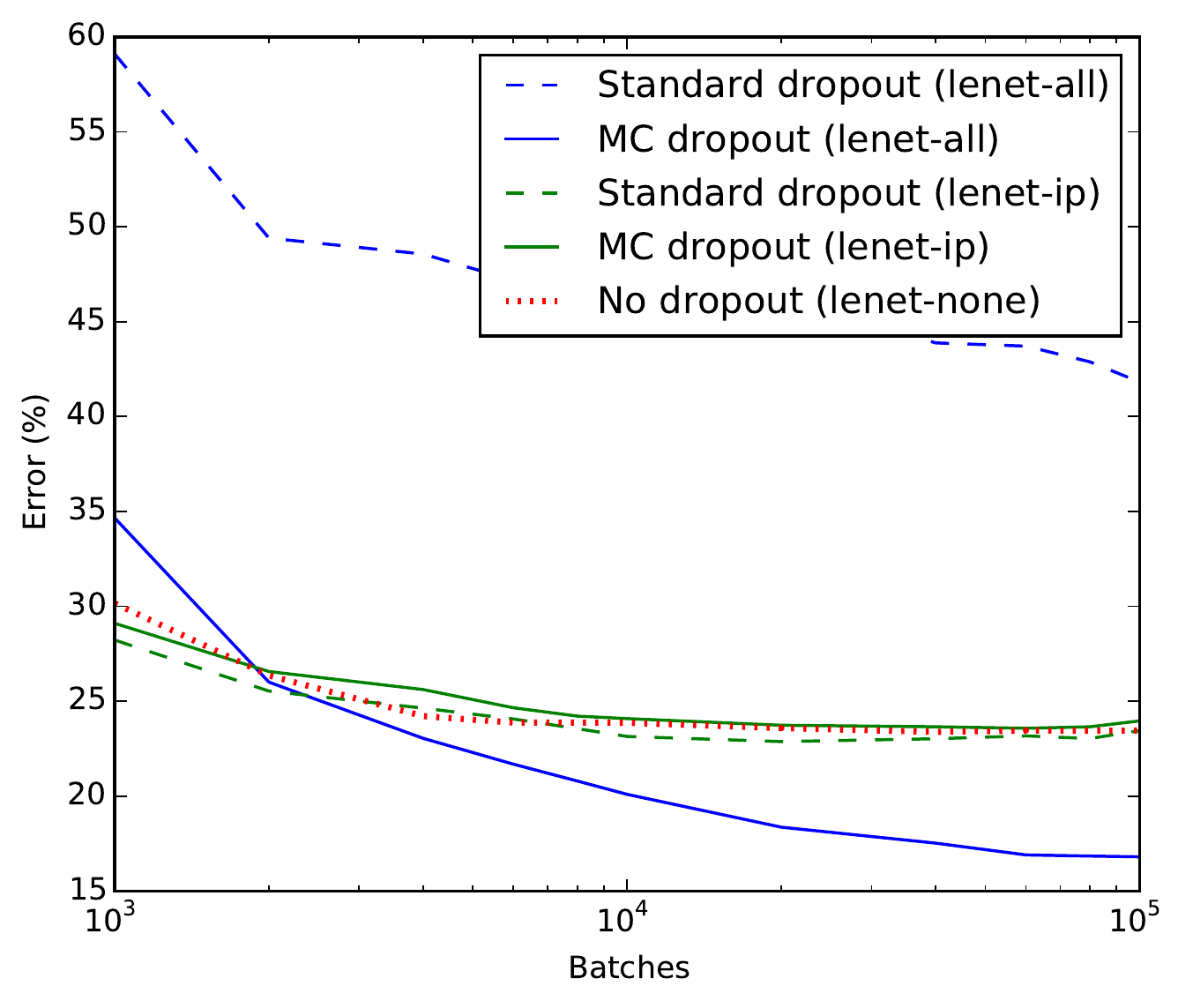}
		\caption{CIFAR-10} \label{fig:bayesian-CNN-cifar}
	\end{subfigure}
	\vspace{1mm}
	\caption{\textbf{Test error 
	for LeNet with dropout applied after every weight layer (lenet-all -- our Bayesian CNN implementation, blue), dropout applied after the fully connected layer alone (lenet-ip, green), and without dropout (lenet-none, dotted red line).} Standard dropout is shown with a dashed line, MC dropout is shown with a solid line. Note that although Standard dropout lenet-all performs very badly on both datasets (dashed blue line), when evaluating \textit{the same network} with MC dropout (solid blue line) the model outperforms all others.} \label{fig:bayesian-CNN}
\end{figure} 

We refer to our Bayesian CNN implementation with dropout used after every parameter layer as ``lenet-all''. We compare this model to a CNN with dropout used after the inner-product layers at the end of the network alone -- the traditional use of dropout in the literature. We refer to this model as ``lenet-ip''. Additionally we compare to LeNet as described originally in \citep{lecun1998gradient} with no dropout at all, referred to as ``lenet-none''. 
We evaluate each dropout network structure (lenet-all and lenet-ip) using two testing techniques. The first is using weight averaging, the standard way dropout is used in the literature (referred to as ``Standard dropout''). This involves multiplying the weights of the $i$'th layer by $p_i$ at test time. 
We use the Caffe \citep{jia2014caffe} reference implementation for this. 
The second testing technique interleaves Bayesian methodology into deep learning. We average $T$ stochastic forward passes through the model following the Bayesian interpretation of dropout derived in eq.\ \eqref{eq:approx_predictive}. 
This technique is referred to here as ``MC dropout''. 
The technique has been motivated in the literature before as model averaging, but never used with CNNs. 
In this experiment we average $T=50$ forward passes through the network.
We stress that the purpose of this experiment is not to achieve state-of-the-art results on either dataset, but rather to compare the different models with different testing techniques.
Full experiment set-up is given in section \ref{sect:appnd:Bayesian Convolutional Neural Networks}.

\citet{krizhevsky2012imagenet} and most existing CNNs literature use Standard dropout after the fully-connected layers alone, equivalent to ``Standard dropout lenet-ip'' in our experiment. \citet[][section 6.1.2]{srivastava2014dropout} use Standard dropout in all CNN layers, equivalent to ``Standard dropout lenet-all'' in our experiment. \citet{srivastava2014dropout} further claim that Standard dropout results in very close results to MC dropout in normal NNs, but have not tested this claim with CNNs. 

\begin{figure}[b!]
\vspace{-5mm}
	\begin{subfigure}[b]{0.5\textwidth}
		\includegraphics[width=\linewidth]{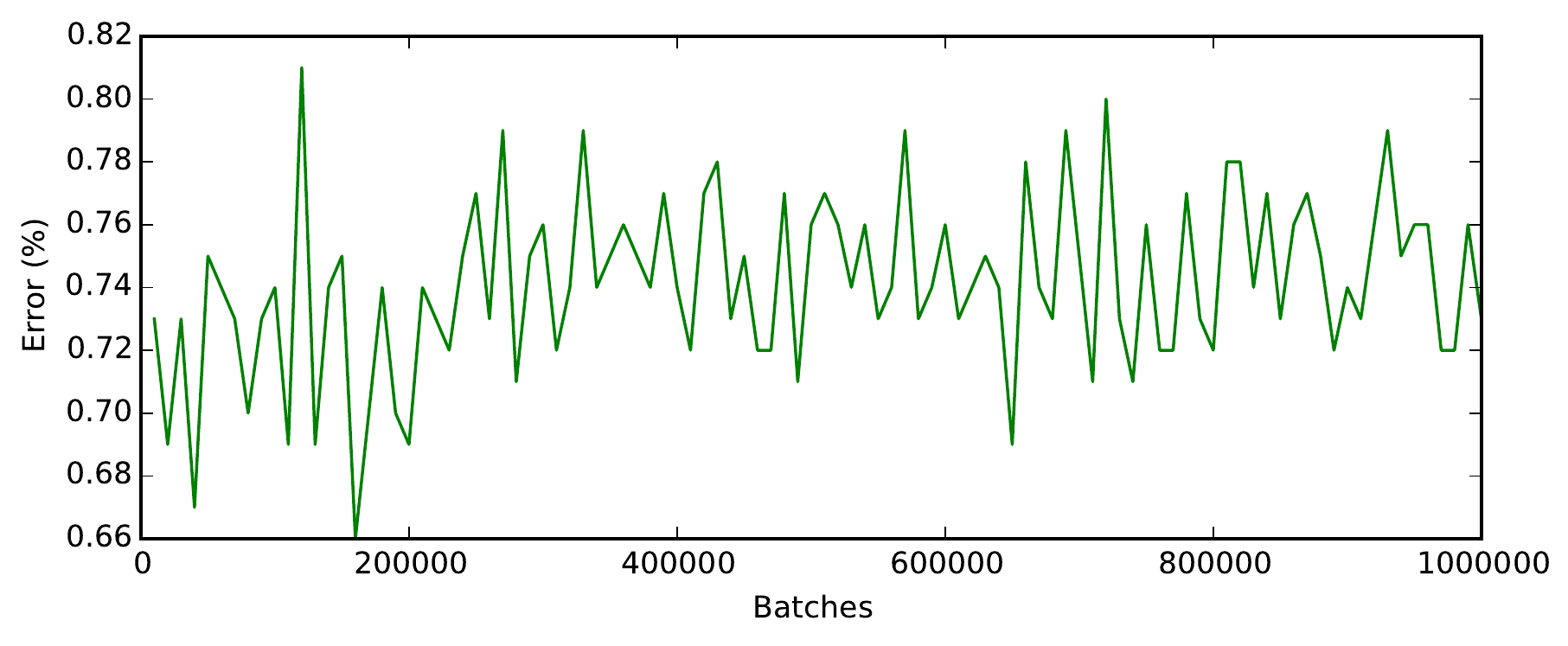}
		\vspace{-6mm}
		\caption{Entire MNIST, Standard dropout + lenet-ip} \label{fig:Standard_dropout_lenet_ip_1}
		\vspace{6mm}
	\end{subfigure}~~~
	\begin{subfigure}[b]{0.5\textwidth}
		\includegraphics[width=\linewidth]{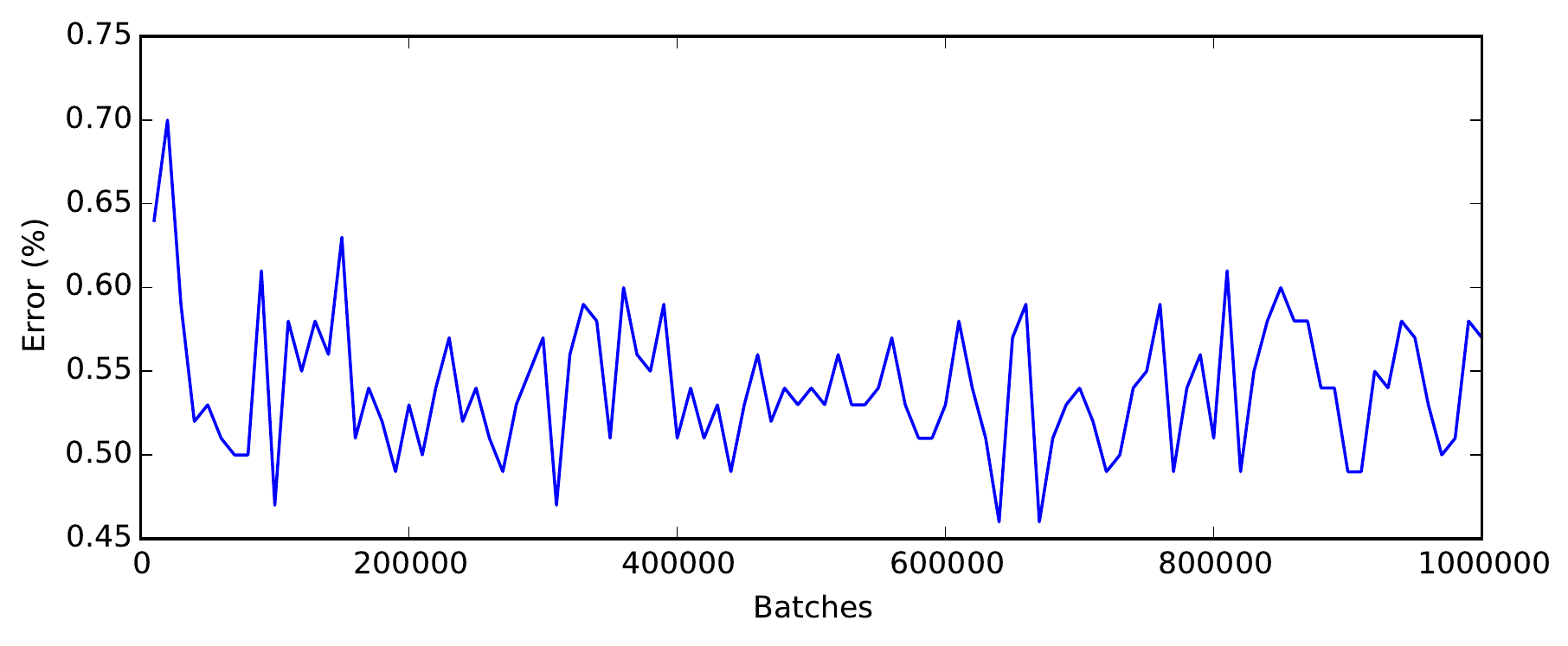}
		\vspace{-6mm}
		\caption{Entire MNIST, MC dropout + lenet-all} \label{fig:MC_dropout_lenet_all_1}
		\vspace{6mm}
	\end{subfigure}
	
	\begin{subfigure}[b]{0.5\textwidth}
		\includegraphics[width=\linewidth]{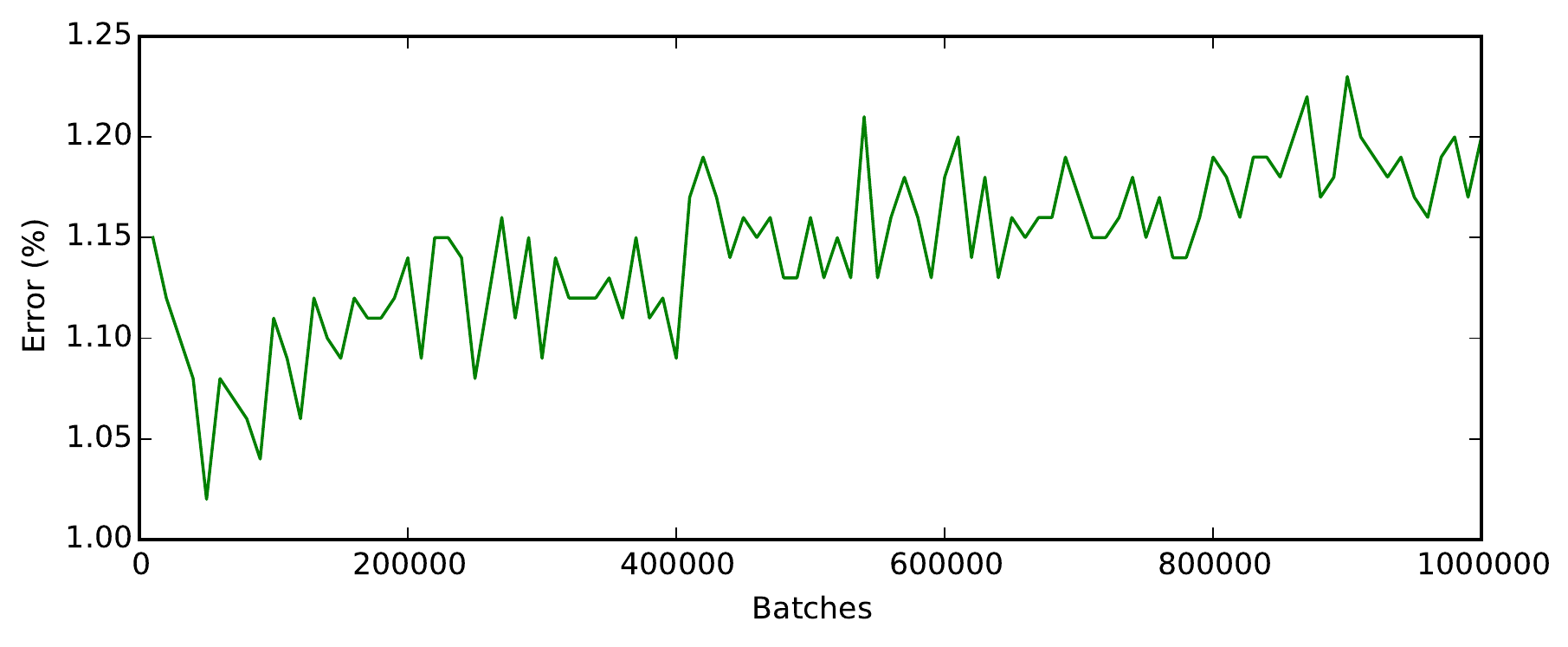}
		\vspace{-6mm}
		\caption{$1/4$ of MNIST, Standard dropout + lenet-ip} \label{fig:Standard_dropout_lenet_ip_4}
		\vspace{6mm}
	\end{subfigure}~~~
	\begin{subfigure}[b]{0.5\textwidth}
		\includegraphics[width=\linewidth]{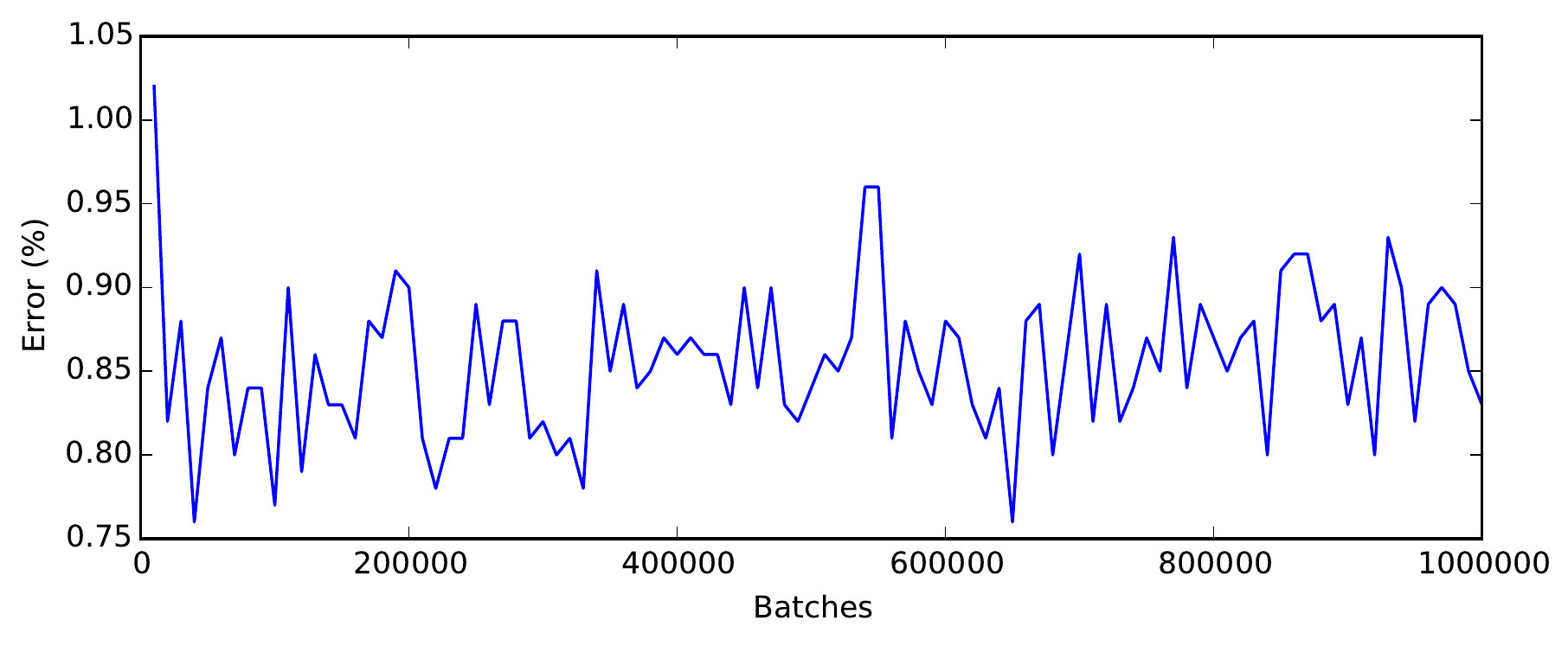}
		\vspace{-6mm}
		\caption{$1/4$ of MNIST, MC dropout + lenet-all} \label{fig:MC_dropout_lenet_all_4}
		\vspace{6mm}
	\end{subfigure}
	
	\begin{subfigure}[b]{0.5\textwidth}
		\includegraphics[width=\linewidth]{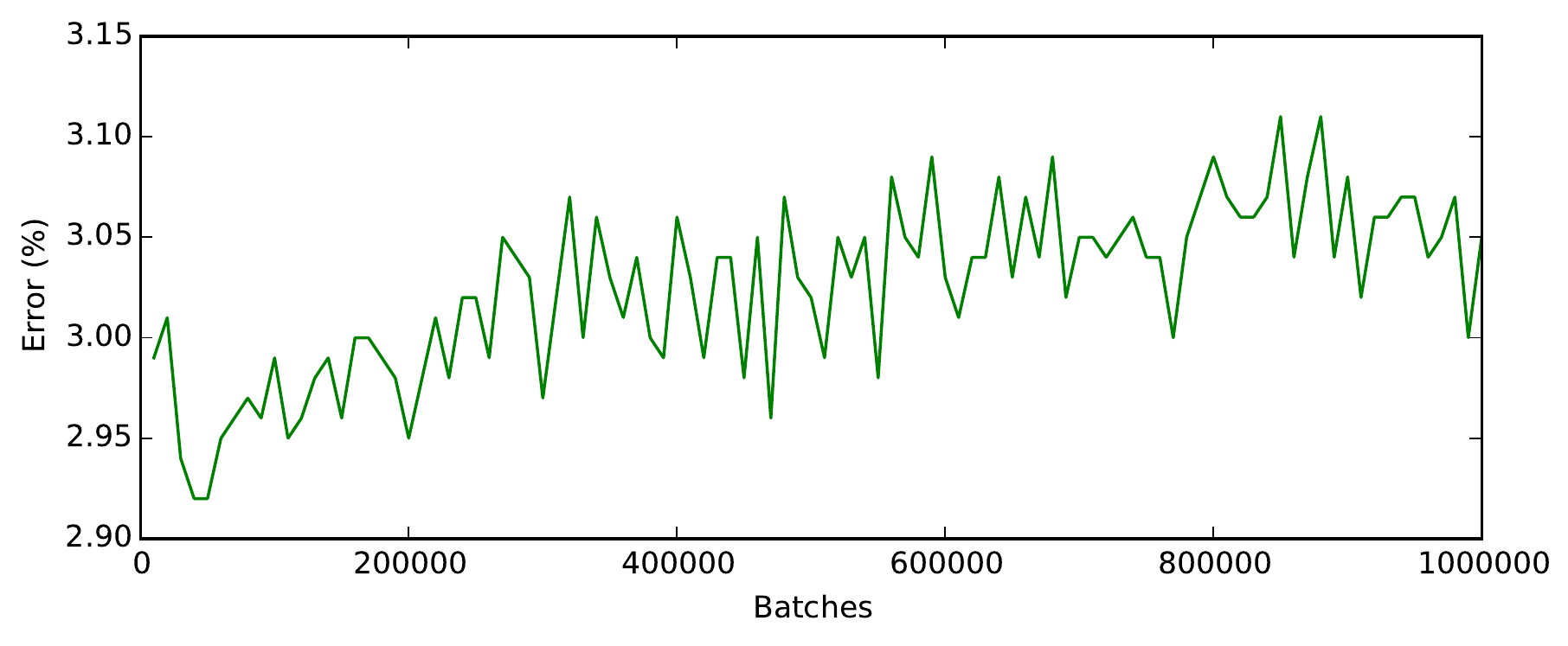}
		\vspace{-6mm}
		\caption{$1/32$ of MNIST, Standard dropout + lenet-ip} \label{fig:Standard_dropout_lenet_ip_32}
	\end{subfigure}~~~
	\begin{subfigure}[b]{0.5\textwidth}
		\includegraphics[width=\linewidth]{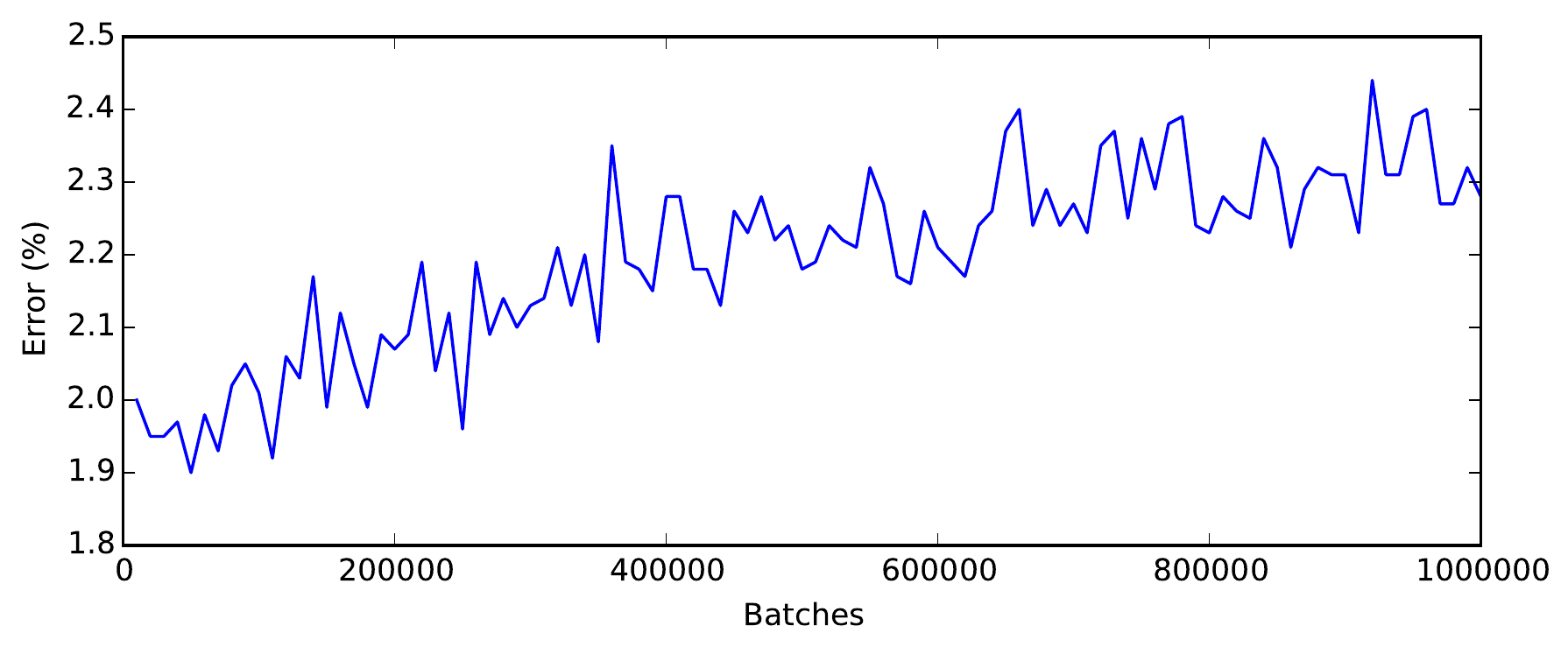}
		\vspace{-6mm}
		\caption{$1/32$ of MNIST, MC dropout + lenet-all} \label{fig:MC_dropout_lenet_all_32}
	\end{subfigure}
	\vspace{1mm}
	\caption{\textbf{Test error of LeNet trained on random subsets of MNIST decreasing in size.} To the left in green are networks with dropout applied after the last layer alone (lenet-ip) and evaluated with Standard dropout (the standard approach in the field), to the right in blue are networks with dropout applied after every weight layer (lenet-all) and evaluated with MC dropout -- our Bayesian CNN implementation. Note how lenet-ip starts over-fitting even with a quarter of the dataset. With a small enough dataset, both models over-fit. MC dropout was used with 10 samples.} \label{fig:mnist-small}
\end{figure}

Figure \ref{fig:bayesian-CNN} shows classification error as a function of batches \textit{on log scale} for all three models (lenet-all, lenet-ip, and lenet-none) with the two different testing techniques (Standard dropout and MC dropout) for MNIST (fig.\ \ref{fig:bayesian-CNN-mnist}) and CIFAR-10 (fig.\ \ref{fig:bayesian-CNN-cifar}).
It seems that Standard dropout in lenet-ip results in improved results compared to lenet-none, with the results more pronounced on the MNIST dataset than CIFAR-10. When Standard dropout testing technique is used with our Bayesian CNN (with dropout applied after every parameter layer -- lenet-all) performance suffers. However by averaging the forward passes of the network the performance of lenet-all supersedes that of all other models (``MC dropout lenet-all'' in both \ref{fig:bayesian-CNN-mnist} and \ref{fig:bayesian-CNN-cifar}). 
Our results suggest that MC dropout should be carried out after all convolution layers.

Dropout has not been used in CNNs after convolution layers in the past, perhaps because empirical results with Standard dropout suggested deteriorated performance (as can also be seen in our experiments). 
Standard dropout approximates model output during test time by weight averaging. 
However the mathematically grounded approach of using dropout at test time is by Monte Carlo averaging of stochastic forward passes through the model (eq.\ \eqref{eq:approx_predictive}). 
The empirical results given in \citet[][section 7.5]{srivastava2014dropout} suggested that Standard dropout is equivalent to MC dropout, and it seems that most research has followed this approximation. However the results we obtained in our experiments suggest that the approximation fails in some model architectures.


\subsection{Model Over-fitting} \label{sect:Model Over-fitting}

We evaluate our model's tendency to over-fit on training sets decreasing in size. We use the same experiment set-up as above, without changing the dropout ratio for smaller datasets. We randomly split the MNIST dataset into smaller training sets of sizes $1/4$ and $1/32$ fractions of the full set. We evaluated our model with MC dropout compared to lenet-ip with Standard dropout -- the standard approach in the field. We did not compare to lenet-none as it is known to over-fit even on the full MNIST dataset.

The results are shown in fig.\ \ref{fig:mnist-small}. For the entire MNIST dataset (figs.\ \ref{fig:Standard_dropout_lenet_ip_1} and \ref{fig:MC_dropout_lenet_all_1}) none of the models seem to over-fit (with lenet-ip performing worse than lenet-all).
It seems that even for a quarter of the MNIST dataset ($15,000$ data points) the Standard dropout technique starts over-fitting (fig.\ \ref{fig:Standard_dropout_lenet_ip_4}). In comparison, our model performs well on this dataset (obtaining better classification accuracy than the best result of Standard dropout on lenet-ip). When using a smaller dataset with $1,875$ training examples it seems that both techniques over-fit, and other forms of regularisation are needed. 

The additional layers of dropout in our Bayesian CNN prevent over-fitting in the model's kernels. This can be seen as a full Bayesian treatment of the model, approximated with MC integration. The stochastic optimisation objective converges to the same limit as the full Bayesian model \citep{blei2012variational, kingma2013auto, rezende2014stochastic, titsias2014doubly, hoffman2013stochastic}. Thus the approximate model possesses the same robustness to over-fitting properties as the full Bayesian model -- approximately integrating over the CNN kernels. 
The Bernoulli approximating variational distribution is a fairly weak approximation however -- a trade-off which allows us to use no additional model parameters. This explains the over-fitting observed with small enough datasets.

%
%
%

\subsection{MC Dropout in Standard Convolutional Neural Networks} \label{sect:MC Dropout in Standard Convolutional Neural Networks}

\begin{table}[t!]
\vspace{-5mm}
	\center
	\def\arraystretch{1.25}
	\begin{tabular}{ccc}
	\multicolumn{1}{c}{} & \multicolumn{2}{c}{\small \textbf{CIFAR Test Error (and Std.)}} \\ 
	\textbf{Model} & \textbf{Standard Dropout} & \textbf{MC Dropout} \\ 
	\hline 
	NIN & $10.43$ & $\mathbf{10.27 \pm 0.05}$ \\ 
	DSN & $9.37$ & $\mathbf{9.32 \pm 0.02}$ \\ 
	Augmented-DSN & $7.95$ & $\mathbf{7.71 \pm 0.09}$ \\ 
	\hline 
	\end{tabular} 
	\caption{\textbf{Test error on CIFAR-10 with the same networks evaluated using Standard dropout versus MC dropout} ($T=100$, averaged with 5 repetitions and given with standard deviation). MC dropout achieves consistent improvement in test error compared to Standard dropout. The lowest error obtained is $7.51$ for Augmented-DSN. 
	} \label{table:cifar}
\end{table}

We evaluate the use of Standard dropout compared to MC dropout on existing CNN models previously published in the literature\footnote{Using 
\href{http://rodrigob.github.io/are_we_there_yet/build/classification_datasets_results.html\#43494641522d3130}{
\texttt{http://rodrigob.github.io/are\_we\_there\_yet/build/\\
classification\_datasets\_results.html\#43494641522d3130}} as a reference.}. The recent state-of-the-art CNN models use dropout after fully-connected layers that are followed by other convolution layers, suggesting that improved performance could be obtained with MC dropout. 

We evaluate two well known models that have achieved state-of-the-art results on CIFAR-10 in the past several years. The first is Network in network (NIN) \citep{lin2013network}. 
The model was extended by \citep{lee2014deeply} who added multiple loss functions after some of the layers -- in effect encouraging the bottom layers to explain the data better. The new model was named a Deeply supervised network (DSN). The same idea was used in \citep{szegedy2014going} to achieve state-of-the-art results on ImageNet.

We assess these models on the CIFAR-10 dataset, as well as on an augmented version of the dataset for the DSN model \citep{lee2014deeply}. We replicate the experiment set-up as it appears in the original papers, and evaluate the models' test error using Standard dropot as well as using MC dropout, averaging $T=100$ forward passes. MC dropout testing gives us a noisy estimate, with potentially different test results over different runs. 
To get faithful results one would need to repeat each experiment several times to get a mean and standard deviation for the test error (whereas standard techniques in the field would usually report the lowest error alone). 
We therefore repeat the experiment 5 times and report the average test error. We use the models obtained when optimisation is done (using no early stopping). We report standard deviation to see if the improvement is statistically significant.

Test error using both Standard dropout and MC dropout for the models (NIN, DSN, and Augmented-DSN on the augmented dataset) are shown in table \ref{table:cifar}. 
As can be seen, using MC dropout a statistically significant improvement can be obtained for all three models (NIN, DSN, and Augmented-DSN), with the largest increase for Augmented-DSN. It is also interesting to note that the lowest test error we obtained for Augmented-DSN (in the 5 experiment repetitions) is 7.51. Our results suggest that MC dropout might improve performance even with standard CNN models.

It is interesting to note that we observed no improvement on ImageNet \citep{deng2009imagenet} using the same models. This might be because of the large number of parameters in the models above compared to the relatively smaller CIFAR-10 dataset size. 
We speculate that our approach offers better regularisation in this setting.
ImageNet dataset size is much larger, perhaps offering sufficient regularisation. 
However labelled data is hard to collect, and in some applications larger amounts of data are not available.
It would be interesting to see if a subset of the ImageNet data could be used to obtain the same results obtained with the full ImageNet dataset with the stronger regularisation suggested in this work. We leave this question for future research.

\subsection{MC Estimate Convergence} \label{sect:MC Estimate Convergence}

\begin{figure}[t!]
\includegraphics[width=\linewidth]{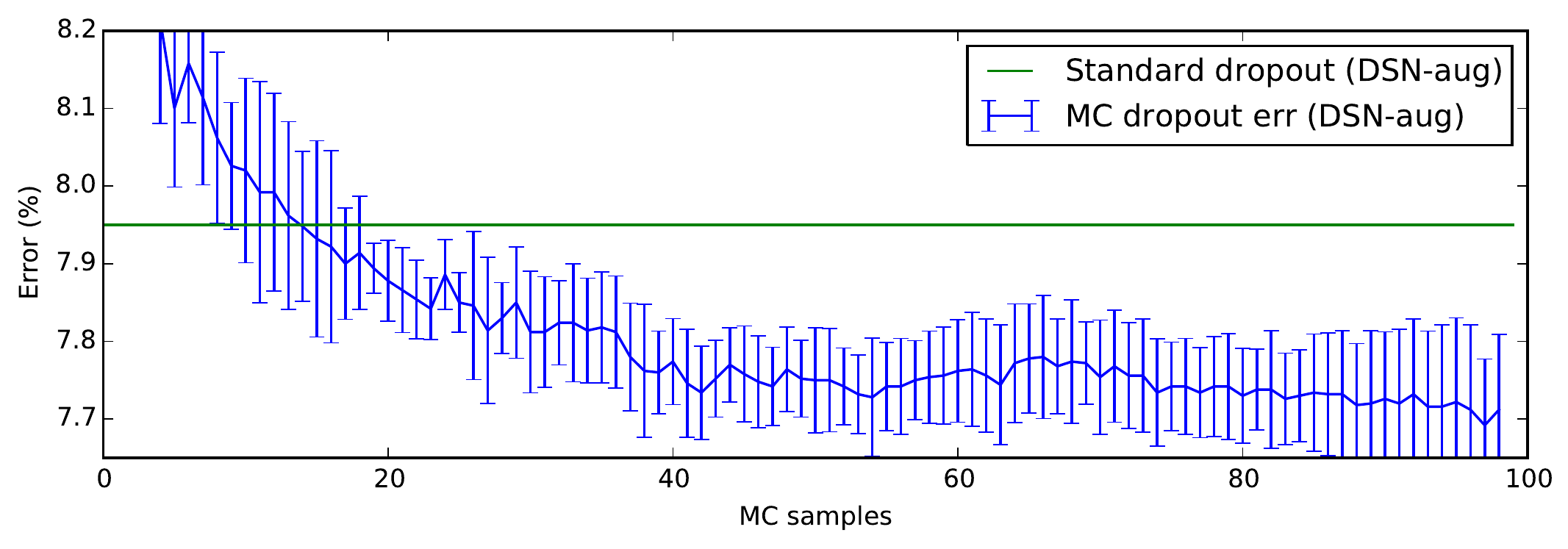}
\caption{\textbf{Augmented-DSN test error for different number of averaged forward passes in MC dropout} (blue) averaged with 5 repetitions, shown with 1 standard deviation. In green is test error with Standard dropout. MC dropout achieves a significant improvement (more than 1 standard deviation) after 20 samples.}  \label{fig:mc-T}
\end{figure}

Lastly, we assess the usefulness of the proposed method in practice for applications in which efficiency during test time is important.
We give empirical results suggesting that 20 samples are enough to improve performance on some datasets. We evaluated the last model (Augmented-DSN) with MC dropout for $T = 1, ..., 100$. We repeat the experiment 5 times and average the results.
In fig.\ \ref{fig:mc-T} we see that within 20 samples the error is reduced by more than one standard deviation. Within 100 samples the error converges to $7.71$. 

This replicates the experiment in \citep[][section 7.5]{srivastava2014dropout}, here with the augmented CIFAR-10 dataset and the DSN CNN model, but compared to \citep[][section 7.5]{srivastava2014dropout} we showed that a significant reduction in test error can be achieved. This might be because CNNs exhibit different characteristics from standard NNs. We speculate that the non-linear pooling layer affects the dropout approximation considerably. 

\section{Conclusions and Future Research}

CNNs work well on large datasets. But labelled data is hard to collect, and in some applications larger amounts of data are not available. The problem then is how to use CNNs with small data -- as CNNs are known to overfit quickly. 
This is because even though dropout is effective in inner-product layers, when it is placed over kernels it leads to diminished results. To solve this we have presented an efficient Bayesian convolutional neural network, offering better robustness to over-fitting on small data by placing a probability distribution over the CNN's kernels. The model's intractable posterior was approximated with Bernoulli variational distributions, requiring no additional model parameters. 
The model implementation uses existing tools in the fields and requires almost no overheads.

Following our theoretical developments casting dropout training as approximate inference in a Bayesian NN, theoretical justification was given for the use of MC dropout as approximate integration of the kernels in a CNN. 
Empirically, we observed that MC dropout improves performance in model architectures for which the standard dropout approximation fails. This comes with a cost of slower test time (as discussed next), therefore optimal choice of inference approximation should be problem dependent.

It is worth noting that the training time of our model is identical to that of existing models in the field, but test time is scaled by the number of averaged forward passes. This should not be of real concern as the forward passes can be done concurrently.
This is explained in more detail in section \ref{sect:appnd:Test Time Complexity} in the appendix.
Future research includes the study of the Gaussian process interpretation of convolution and pooling. These might relate to existing literature on convolutional kernel networks \citep{mairal2014convolutional}. Furthermore, it would be interesting to see if a subset of the ImageNet data could be used to obtain the same results with the stronger regularisation suggested in this work. We further aim to study how the learnt filters are affected by dropout with different probabilities.

\section*{Acknowledgements} 
The authors would like to thank Mr Alex Kendall, Mr Eunbyung Park, and the reviewers for their helpful comments. Yarin Gal is supported by the Google European Fellowship in Machine Learning.

\nocite{langley00}

{\small
\bibliography{example_paper}
\bibliographystyle{unsrtnat}
}

\newpage
\appendix
\section{Experiment Set-up}
\subsection{Bayesian Convolutional Neural Networks}\label{sect:appnd:Bayesian Convolutional Neural Networks}
For MNIST we use the LeNet network as described in \citep{lecun1998gradient} with dropout probability $0.5$ in every dropout layer.
The model used with CIFAR-10 is set up in an identical way, with the only difference being the use of 192 outputs in each convolution layer instead of 20 and 50, as well as 1000 units in the last inner product layer instead of 500. 

We ran a stochastic gradient descent optimiser for $1e7$ iterations for all MNIST models and $1e5$ iterations for all CIFAR-10 models. We used learning rate policy $\text{base-lr} * (1 + \gamma * \text{iter})^{-p}$ with $\gamma=0.0001, p=0.75$, and momentum $0.9$. We used base learning rate $0.01$ and weight decay $0.0005$.
All models where optimised with the same parameter settings. 

\subsection{Test Time Complexity}\label{sect:appnd:Test Time Complexity}

Our improved results come with a potential price: longer test time. 
The training time of our model is identical to that of existing models in the field. The test time is scaled by $T$ -- the number of averaged forward passes through the network.
However this should not be of real concern in real world applications, as CNNs are often implemented on distributed hardware. This allows us to obtain MC dropout estimates in constant time almost trivially. This could be done by transferring an input to a GPU and setting a mini-batch composed of the same input multiple times. In dropout we sample different Bernoulli realisations for each output unit and each mini-batch input, which results in a matrix of probabilities. Each row in the matrix is the output of the dropout network on the same input generated with different random variable realisations. Averaging over the rows results in the MC dropout estimate.
Further, many models are tested with multiple crops of the same input. This could be done with stochastic forward passes instead of averaged weights.

\end{document}